\documentclass[conference]{IEEEtran}
\IEEEoverridecommandlockouts
\usepackage{cite}
\usepackage{amsmath,amssymb,amsfonts}
\usepackage{algorithmic}
\usepackage{graphicx}
\usepackage{textcomp}
\usepackage{xcolor}
\usepackage{url}

\def\BibTeX{{\rm B\kern-.05em{\sc i\kern-.025em b}\kern-.08em
    T\kern-.1667em\lower.7ex\hbox{E}\kern-.125emX}}
\begin{document}

\title{A Cascaded Neural Network System For Rating Student Performance In Surgical Knot Tying Simulation\\
}

\author{\IEEEauthorblockN{1\textsuperscript{st} Yunzhe Xue}
\IEEEauthorblockA{\textit{Department of Data Science} \\
\textit{New Jersey Institute of Technology}\\
Newark, NJ, USA \\
yx277@njit.edu}
\\
\IEEEauthorblockN{4\textsuperscript{th} Nell M. Patel}
\IEEEauthorblockA{\textit{Colorectal Surgery Division} \\
\textit{Robert Wood Johnson Hospital}\\
New Brunswick, NJ, USA \\
malonene@rwjms.rutgers.edu}
\and
\IEEEauthorblockN{2\textsuperscript{nd} Olanrewaju Eletta}
\IEEEauthorblockA{\textit{General Surgery} \\
\textit{Robert Wood Johnson Hospital}\\
New Brunswick, NJ, USA \\
oe78@rwjms.rutgers.edu}
\\
\IEEEauthorblockN{5\textsuperscript{th} Advaith Bongu}
\IEEEauthorblockA{\textit{Transplant Surgery Division} \\
\textit{Robert Wood Johnson Hospital}\\
New Brunswick, NJ, USA \\
Advaith.Bongu@rwjbh.org}
\and
\IEEEauthorblockN{3\textsuperscript{rd} Justin W. Ady}
\IEEEauthorblockA{\textit{Vascular and Endovascular Surgery} \\
\textit{Robert Wood Johnson Hospital}\\
New Brunswick, NJ, USA \\
jwa60@rwjms.rutgers.edu}
\\
\IEEEauthorblockN{6\textsuperscript{th} Usman Roshan}
\IEEEauthorblockA{\textit{Department of Data Science} \\
\textit{New Jersey Institute of Technology}\\
Newark, NJ, USA \\
usman@njit.edu}
}

\maketitle

\begin{abstract}
As part of their training all medical students and residents have to pass basic surgical tasks such as knot tying, needle-passing, and suturing. Their assessment is typically performed in the operating room by surgical faculty where mistakes and failure by the student increases the operation time and cost. This evaluation is quantitative and has a low margin of error. Simulation has emerged as a cost effective option but it lacks assessment or requires additional expensive hardware for evaluation. Apps that provide training videos on surgical knot trying are available to students but none have evaluation. We propose a cascaded neural network architecture that evaluates a student's performance just from a video of themselves simulating a surgical knot tying task. Our model converts video frame images into feature vectors with a pre-trained deep convolutional network and then models the sequence of frames with a temporal network. We obtained videos of medical students and residents from the Robert Wood Johnson Hospital performing knot tying on a standardized simulation kit. We manually annotated each video and proceeded to do a five-fold cross-validation study on them. Our model achieves a median precision, recall, and F1-score of 0.71, 0.66, and 0.65 respectively in determining the level of knot related tasks of tying and pushing the knot. Our mean precision score averaged across different probability thresholds is 0.8. Both our F1-score and mean precision score are 8\% and 30\% higher than that of a recently published study for the same problem. We expect the accuracy of our model to further increase as we add more training videos to the model thus making it a practical solution that students can use to evaluate themselves.
\end{abstract}


\section{Introduction}
Surgical skill training and evaluation has a high cost in terms of impacting operating room efficiency \cite{bridges1999financial,allen2016effect}. Simulation has emerged as an effective method for basic technical skills acquisition. We see that surgical training skills acquired in simulation transfer to the operating room and also shorten learning curves \cite{palter2011individualized,dawe2014systematic}. Simulation can decrease inexperience and translate to improved efficiency and patient safety outcomes \cite{zendejas2011simulation,cox2015moving}. %An estimated 5-17\% of trainees have an innate technical ability while others may take more time to achieve competency \cite{louridas2017practice}. 

The primary limitation of simulation is automatic and accurate feedback to the student on their performance. Evaluation of a task in simulation requires resources and time \cite{lu2021simulation,nicholas2019simulation}. Simulators can be made low cost \cite{sharma2020low} but automatic evaluation is still lacking. Simulation with expert level feedback translates to improved operating room performance \cite{price2011randomized}, but without an automatic feedback component their usage is limited. A self-contained simulation complete with feedback facilitates repetitive use that prevents skills degradation \cite{stefanidis2015simulation}. 

As a step towards making simulation more broadly available to students we present an artificial intelligence system that students can use to evaluate themselves in surgical knot-tying - a basic yet fundamentally important task in surgical training. We present a cascaded neural network system trained on videos of students and residents performing knot tying in a standardized simulated environment. In cross-validation studies we show that our model achieves a median precision, recall, and F1-score of 0.71, 0.66, and 0.65 respectively. Our mean precision score averaged across different probability thresholds (versus the standard 0.5 probability threshold) is 0.8. Compared to a recently published study \cite{nagaraj2022developing} on the same problem our model has an 8\% higher F1-score and a 30\% higher mean precision score. Our model also gives a frame by frame analysis of the video that allows detailed feedback on areas where the student can focus on improvement. In the rest of the paper we present our data, model, and detailed results.

\section{Methods}

\subsection{Data}
\begin{figure}[t!]
\centerline{\includegraphics[scale=0.9]{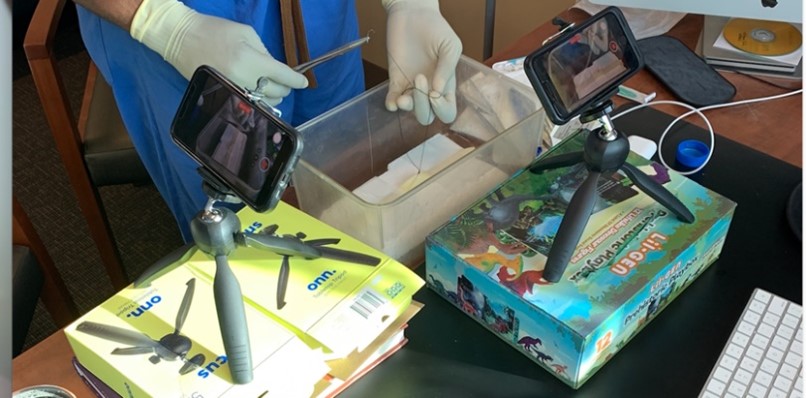}}
\caption{Simulation kit recording the student's exercise. In this study we only use videos showing the horizontal view. A more sophisticated approach could utilize both horizontal and lateral views as recorded by separate cameras.}
\label{kit}
\end{figure}

We collected videos of 27 medical students and 7 residents performing surgical knot tying in simulation under Institution Review Board (IRB) approval. We have a standardized kit for knot tying as shown in Figure~\ref{kit}. For a given video our medical team annotated portions of it into one of four action categories: \{Waiting, Needling, Pushing knot, Tying knot\} and one of three level categories: \{Good, OK, Bad\}. This gives us a total of 12 classes making it a multiclass classification problem. 

\begin{figure}[h!]
\begin{tabular}{c}
\centerline{\includegraphics[scale=0.35]{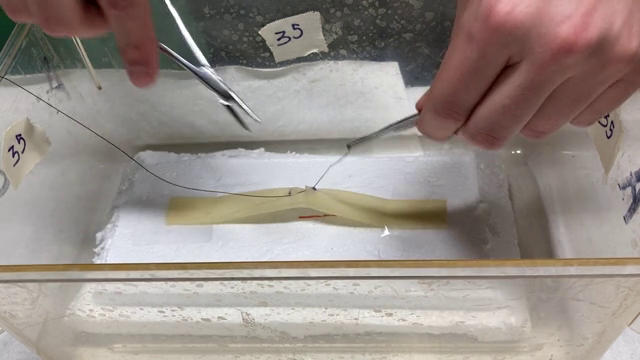}} \\ 
(a) Needling \\
\centerline{\includegraphics[scale=0.35]{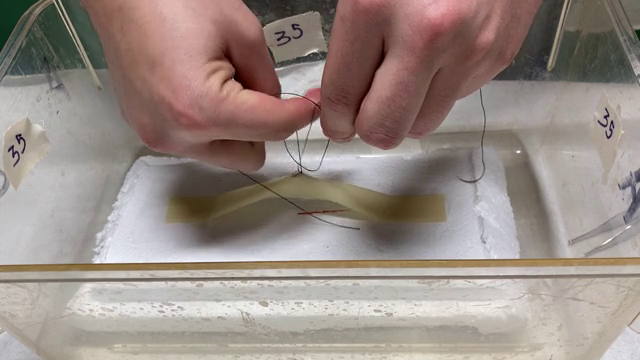}}  \\
(b) Tying knot \\
\centerline{\includegraphics[scale=0.35]{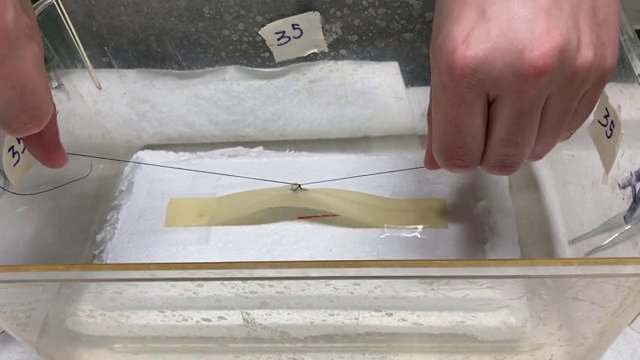}}  \\
(c) Pushing knot \\
\end{tabular}
\caption{Fourth year student needling, tying knot, and pushing knot. We see pulling and fumbling in all three actions.}
\label{student}
\end{figure}

\begin{figure}[h!]
\begin{tabular}{c}
\centerline{\includegraphics[scale=0.35]{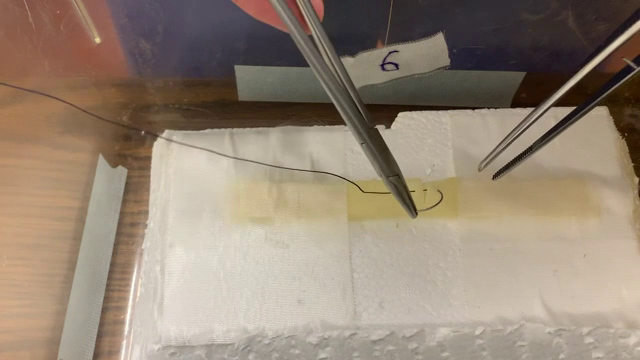}} \\ 
(a) Needling \\
\centerline{\includegraphics[scale=0.35]{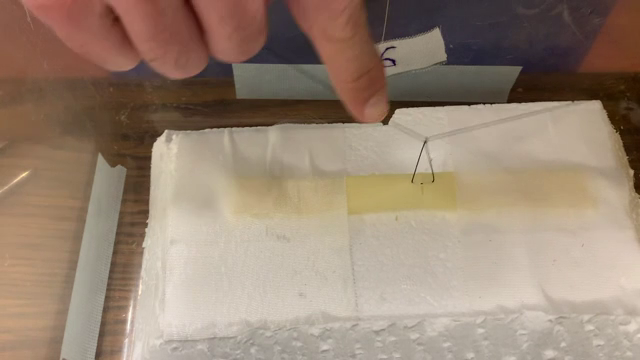}}  \\
(b) Tying knot \\
\centerline{\includegraphics[scale=0.35]{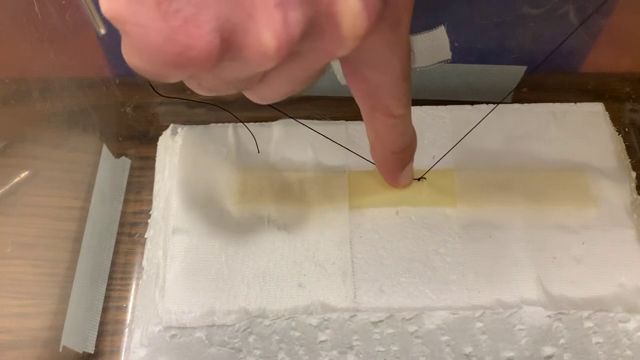}}  \\
(c) Pushing knot \\
\end{tabular}
\caption{Fifth year resident needling, tying knot, and pushing knot. There is no pulling, and we see that both tying and pushing the knot are correctly performed.}
\label{expert}
\end{figure}

In Figures~\ref{student} and~\ref{expert} we show examples of a fourth year medical student and a fifth year medical resident performing needling, pushing knot, and tying in our simulation environment. We can see that the medical student has considerably more pull on the fake skin than the expert. We also see that the student fumbles during tying and instead of pushing on the knot they pull on it. During our annotation we labeled the student's examples in Figure~\ref{student}(a) through (c) as Needling-Bad, Tying knot-Bad, and Pushing knot-Bad respectively. For the expert, however, we labeled them with the same action but a level of good. Each combination represents a unique category.

Each video has 29 to 30 frames (images) per second. The mean length of our videos is 99 seconds with a standard deviation of 36.5, and the mean number of frames in each video is 2954 with standard deviation of 1095. In Table~\ref{videos} we show the student level, number of frames, and length in seconds of each video.

\begin{table}[htbp]
\caption{Number of frames (images) and length in seconds of videos used in our study. MS denotes medical student and the number denotes their year of study. PGY denotes a medical resident in training.}
\begin{center}
\begin{tabular}{|c|c|c|}
\hline
\textbf{Student level}&\textbf{Number of frames}&\textbf{Video length (seconds)} \\
MS3 &	2353&	79\\
MS3	 & 3433&	115\\
MS4	& 3976&	133\\
MS4	& 2241&	75\\
MS4	& 2891&	97\\
MS4	&5785	&193\\
MS4	&2270	&76\\
MS4&	3670&	123\\
MS4&	2494&	84\\
MS4&	2941&	99\\
MS4&	3990&	134\\
MS4&	6006&	201\\
MS4&	4183&	140\\
MS4&	2276&	76\\
MS4&	2657&	89\\
MS4&	3923&	131\\
MS4&	3656&	122\\
MS4&	2534&	85\\
MS4&	2716&	91\\
MS4&	2621&	88\\
MS4&	2356&	79\\
MS4&	3502&	117\\
MS4&	2303&	77\\
MS4&	1888&	63\\
MS4&	3177&	106\\
MS4&	4293&	144\\
MS4&	3340&	112\\
MS4&	2863&	96\\
PGY1&	3303&	111\\
PGY4&	1489&	50\\
PGY4&	1538&	52\\
PGY4&	1802&	61\\
PGY5&	1286&	43\\
PGY5&	1965&	66\\
PGY5&	1680&	57\\ \hline
\end{tabular}
\label{videos}
\end{center}
\end{table} 

\subsection{Models}
\begin{figure}[htbp!]
%\centerline{\includegraphics[width=.7\linewidth]{model_betterpicture2}} 
\centerline{\includegraphics[scale=.28]{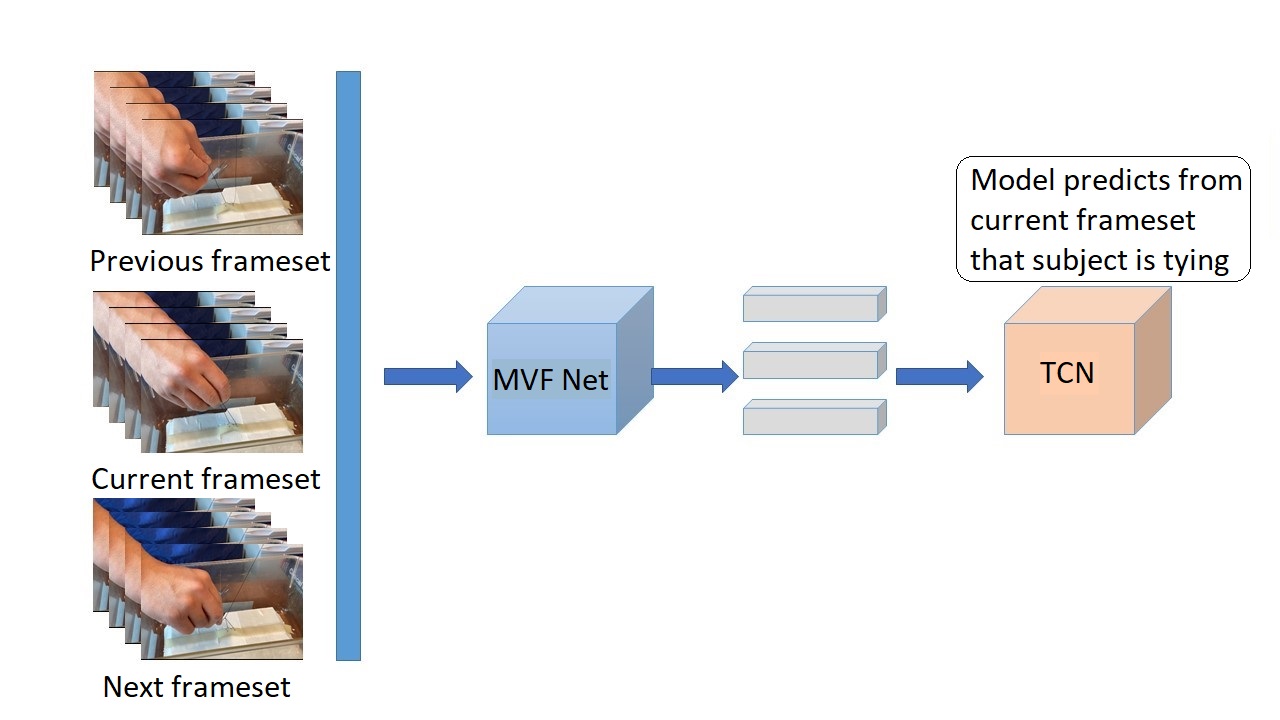}} 
\caption{Our high-level model design. We perform end-to-end training of our model.}
\label{model}
\end{figure}

We propose a cascaded system of two neural networks: MVFNet for feature extraction and a temporal network for modeling time dependency between the feature vectors. During training we give the model several frames of the video and the target prediction as the action and level category of the middle one. For example if the person is tying without fumbling and pulling then the output category would be Tying knot-Good - this is one of the 12 classes our model outputs. We illustrate this in Figure~\ref{model} where we show our overall model architecture.

\subsubsection{MVFNet}
\begin{figure*}[htbp!]
%\centerline{\includegraphics[scale=0.25]{MVF-ResNet50}} 
\centerline{\includegraphics[width=.9\linewidth]{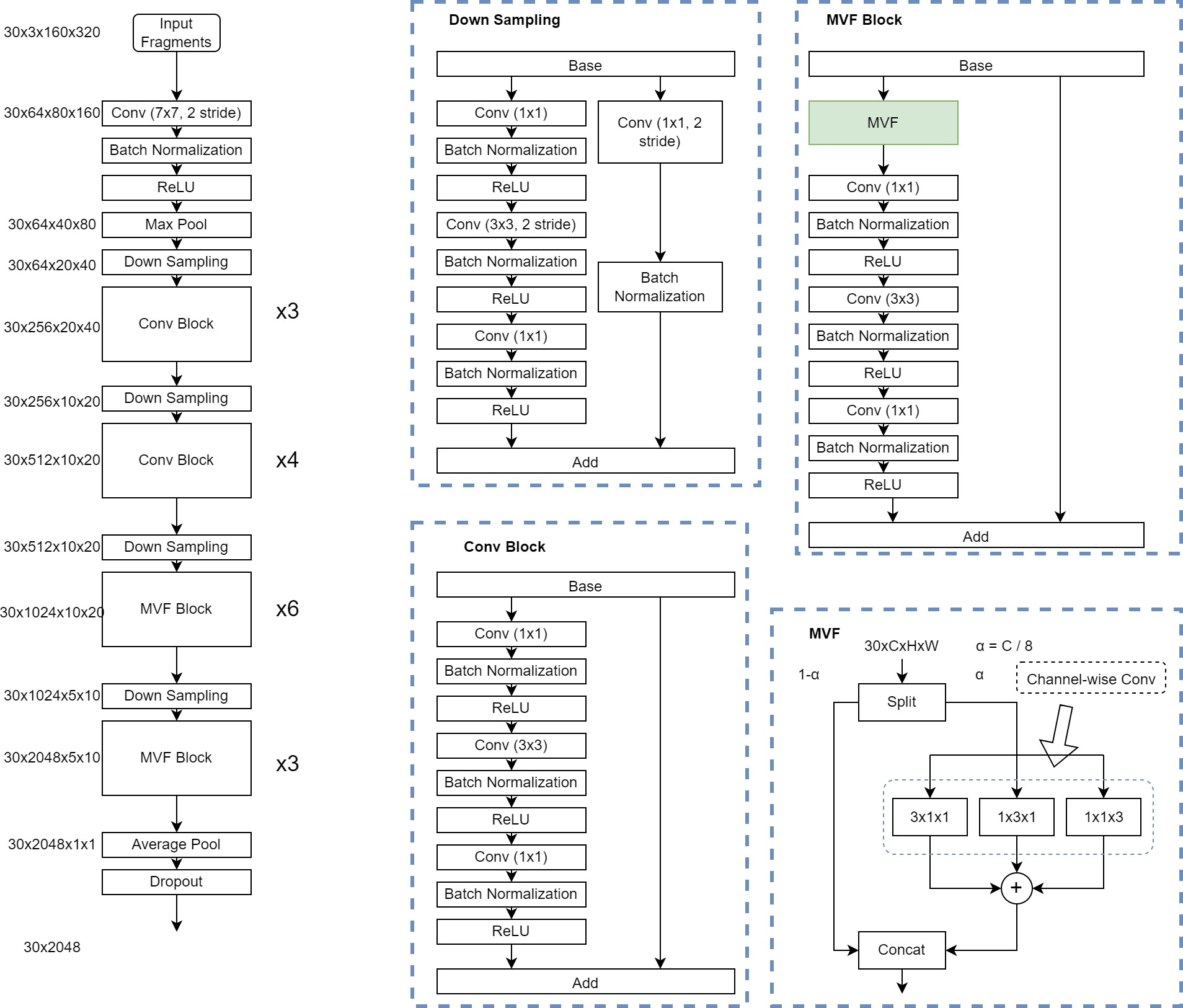}} 
\caption{MultiView Fusion Network (MVFNet) model \cite{wu2021mvfnet}}
\label{mvfnet}
\end{figure*}

The first model in our system is called MultiView Fusion Network (MNVFNet) \cite{wu2021mvfnet} whose architecture we show in Figure~\ref{mvfnet}. This uses 3D convolutions which perform convolution on a set of frames of a video (that we refer to as a \emph{frameset}) represented as a sequence of images in a 3D matrix. This model was trained on the Kinetics video dataset \cite{kay2017kinetics} of 400 human actions containing at least 400 videos for each action. Its pre-trained version is available from the author Github site here \url{https://github.com/whwu95/MVFNet}. The advantage of a pre-trained model is that has been trained on a large set of videos already and requires only fine-tuning specific to our training dataset. The output of MVFNet is a single feature vector from an input of a set of images. 

%\begin{figure}[htbp!]

\subsubsection{Temporal convolutional network}
\begin{figure*}[htbp!]
\centerline{\includegraphics[width=.9\linewidth]{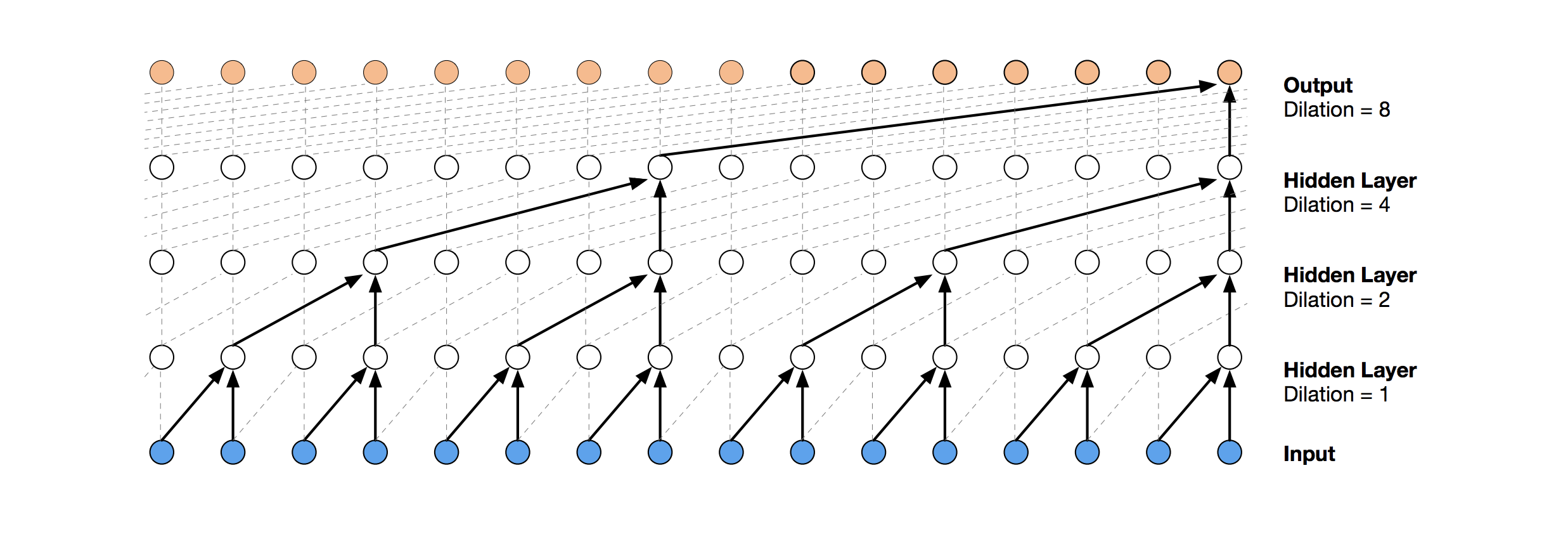}} 
\caption{Dilated convolutions used in our temporal network \cite{oord2016wavenet}}
\label{tcn}
\end{figure*}

We give the sequence of feature vectors from MVFNet as input to a temporal convolutional network. Temporal convolutions were first introduced as dilated causal convolutions shown in Figure~\ref{tcn} \cite{oord2016wavenet}. They are easy to train and have shown to be accurate compared to traditional recurrent neural networks \cite{bai2018empirical}. The output of this network is the action-level designation of the middle feature vector in the input.  

\subsubsection{Ensembling}
Deep neural network models like the one we are using in this study can be sensitive to the choice of seed for the random number generator. Previous work shows that the choice of seed can affect the accuracy of the model \cite{picard2021torch}, but at the same time different seeds can be used to produce an ensemble that is likely to be more accurate and robust \cite{bethard2022we}. Thus, instead of training a single model we train 10 models each with a different initial seed for the random number generator. Below when we refer to our model we refer to the ensemble where we take the majority vote output when making a classification. 

\section{Results}

\subsection{Five-fold cross-validation results}
We perform a standard five-fold cross-validation experiment to evaluate our model's accuracy. We divide our videos into five subsets $\{S_0, S_1, S_2, S_3, S_4\}$ and train models on all combinations of four of the five subsets: $\{S_0, S_1, S_2, S_3\}$, $\{S_0, S_1, S_2, S_4\}$, $\{S_0, S_1, S_3, S_4\}$, $\{S_0, S_2, S_3, S_4\}$, and $\{S_1, S_2, S_3, S_4\}$. This gives us five models that we evaluate on the fifth subset omitted from training. The precision, sensitivity, and F1-score are popular metrics for measuring the binary classification performance of a model. Roughly speaking, precision measures the accuracy of retrieving positives (or class of interest) - if a classifier predicted everything as belonging to positive class it would have precision of 100\%. The sensitivity measures the accuracy of making a mistake in predicting positives - this would be 0\% if the model predicted everything as positive. The F1-score is the harmonic mean of precision and sensitivity and loosely speaking corrects for imbalance in predictions.

These are well-defined for binary classification but in our case we have a multi-class classification in each frameset. A simple extension to multiclass case is to calculate the metrics for each class at a time in a one-vs-all manner \cite{godbole2004discriminative}. For example suppose we have a video of 10 framesets and we want to evaluate metrics for the level of knot tying and knot pushing considered together. We have three classes for the level: 0 that denotes 'good', 1 denotes 'okay', and 2 that denotes 'bad'. In the one-vs-all setting we compute three sets of metrics: class good vs classes okay and bad considered together, class okay vs classes good and bad, and class bad vs classes good and okay. 

We illustrate this with an example of good vs classes okay and bad. Suppose in framesets 1 through 3 and 5 through 7 the ground truth level is 'good' for knot tying and pushing. We denote this as $gt=[2,0,0,0,2,0,0,0,1,1,2]$ where the first item has index 0. To evaluate the metrics for the 'good' level we look at the model's predictions for knot related tasks (tying and pushing) in framesets 1 through 3 and 5 through 7, which is where the ground truth is 'good'. If a prediction is good (0) we consider it a true positive, if the prediction is okay or bad (1 or 2) we consider it a false negative. If the prediction outside these framesets is good we consider that a false positive. So for example if the prediction is $pr=[2,\textcolor{blue}{0,0},\textcolor{red}{1},1,\textcolor{red}{1},\textcolor{blue}{0,0},\textcolor{green}{0},2,2]$ we have $TP=4$ (shown in blue), $FN=2$ (in red), and $FP=1$ (shown in green). With these in hand we can calculate the precision, sensitivity, and F1-scores for class 'good' as shown below. 

%\begin{table}[htbp]
\begin{center}
\begin{tabular}{cc}
$precision=\frac{TP}{TP+FP} $ & $ sensitivity=\frac{TP}{TP+FN}$\\
\\
\multicolumn{2}{c}{$F1score=2\times\frac{precision\times sensitivity}{precision + sensitivity}$} \\
\end{tabular}
\end{center}
%\end{table}

We then calculate the same metrics above for 'okay' and 'bad' and take their averaged weighted by class. For example average weighted F1-score would be $\frac{6}{11}\times F1score(good)+\frac{2}{11}\times F1score(okay)+\frac{3}{11}\times F1score(bad)$. In this way we get the F1-score in the multiclass setting.

We measure the three metrics for each video in our test set. In Table~\ref{metrics} we report the median precision, recall, and F1-score of the level of knot related tasks as well the activity the student is performing. We see that our model has a median precision, sensitivity, and F1-score of 0.71, 0.66, and 0.65 respectively for the level of tying and pushing the knot considered as one action. If we consider metrics for level of tying and pushing the knot as separate actions we see that the F1-score for tying is lower at 0.46 but for pushing the knot the F1-score is 0.65. In both actions separately our model is likely to make more mistakes in determining the level (as indicated by the sensitivity) than if we considered them as one action.

\begin{table}[htbp]
\caption{Precision, sensitivity, and F1-score of our model for the level (good, okay, bad) of tying and pushing knot considered as one action and also considered as separate actions.}
\begin{center}
\begin{tabular}{|c|c|c|c|}
\hline
\textbf{Task}&\textbf{Precision}&\textbf{Sensitivity}&\textbf{F1 score}\\
\hline
\textbf{Tying and pushing }&  & & \\ 
\textbf{knot level (median)}& 0.71  & 0.66 & 0.65 \\ \hline
\textbf{Tying knot}&  & & \\ 
\textbf{level (median)}& 0.75  & 0.44 & 0.46 \\ \hline
\textbf{Pushing knot}&  & & \\ 
\textbf{level (median)}& 0.89  & 0.5 & 0.65 \\ \hline 
\end{tabular}
\label{metrics}
\end{center}
\end{table}

To better understand our metrics we compare them to the same metrics of a recent study on evaluating level of surgical knot tying \cite{nagaraj2022developing}. In that study authors have a training dataset of 229 videos which is more than 6 times the number that we have. That study also evaluates an entire video as pass or fail whereas we look at each frameset of the video. We see in Table~\ref{metrics2} that our mean F1-score is 8\% higher than their reported one. We see that although their precision is high at 0.92, their recall is much lower which indicates that their model is biased towards one of the classes. On other other hand, our precision and sensitivity are closer to the F1-score indicating a balanced prediction.

\begin{table}[htbp]
\caption{We compare our mean metrics to those of a related study for the same problem. Also shown are standard deviations after the $\pm$ symbol.}
\begin{center}
\begin{tabular}{|c|c|c|c|}
\hline
\textbf{Task}&\textbf{Precision}&\textbf{Sensitivity}&\textbf{F1 score}\\
\hline
\textbf{Tying and pushing} &  &  & \\
\textbf{knot level (mean)} & 0.69 $\pm$ 0.33 & 0.62 $\pm$ 0.32 & 0.62 $\pm$ 0.33 \\ \hline
\textbf{Knot level \cite{nagaraj2022developing}}& 0.92 $\pm$ 0.44 & 0.38 $\pm$ 0.31 & 0.54 $\pm$ 0.32 \\
\hline
%\textbf{Task}&\multicolumn{3}{|c|}{\textbf{Table Column Head}} \\
%\cline{2-4} 
%\textbf{Head} & \textbf{\textit{Table column subhead}}& \textbf{\textit{Subhead}}& \textbf{\textit{Subhead}} \\
\end{tabular}
\label{metrics2}
\end{center}
\end{table}

To gain further insight into our metrics we measure the average precision score \cite{zhu2004recall}. This is defined as the weighted mean of precisions at different thresholds with the increase in sensitivity from the last threshold as the weight. In other words, this gives us the model precision for different probability thresholds. The values we reported above for knot related tasks of tying and pushing knot are for a probability threshold of 0.5 since we are doing binary classification in a one-vs-all manner. To measure the effect of different thresholds we average with the mean precision score. We measure the mean precision score for each video in our test set and report the mean value and standard deviation shown after $\pm$ in Table~\ref{meanprecision}. We see that our mean precision score is 31\% higher than the one reported in the related study \cite{nagaraj2022developing}.

\begin{table}[htbp]
\caption{We compare our mean precision score of the level of knot related tasks (tying and pushing) to that of a related study for the same problem.}
\begin{center}
\begin{tabular}{|c|c|}
\hline
\textbf{Task}&\textbf{Mean precision score} \\
\hline
\textbf{Tying and pushing} &   \\
\textbf{knot level (mean)} & 0.8 $\pm$ 0.2  \\ \hline
\textbf{Knot level \cite{nagaraj2022developing}} & 0.49 \\
\hline
%\textbf{Task}&\multicolumn{3}{|c|}{\textbf{Table Column Head}} \\
%\cline{2-4} 
%\textbf{Head} & \textbf{\textit{Table column subhead}}& \textbf{\textit{Subhead}}& \textbf{\textit{Subhead}} \\
\end{tabular}
\label{meanprecision}
\end{center}
\end{table}

In Table~\ref{metrics3} we show our model metrics for just determining the action the student is performing. We see higher scores in all three metrics indicating that knot tying related activity recognition is easier than determining how well the student is performing the activity.

\begin{table}[htbp]
\caption{Our median model metrics for recognizing the action that the student is performing, such as waiting, needling, tying knot, and pushing knot.}
\begin{center}
\begin{tabular}{|c|c|c|c|}
\hline
\textbf{Task}&\textbf{Precision}&\textbf{Sensitivity}&\textbf{F1 score}\\
\hline
\textbf{Action recognition}& 0.86 & 0.82 & 0.82 \\
%copy& More table copy$^{\mathrm{a}}$& &  \\
\hline
%\multicolumn{4}{l}{$^{\mathrm{a}}$Sample of a Table footnote.}
\end{tabular}
\label{metrics3}
\end{center}
\end{table}

Finally we compare the metrics of our ensemble model to one of the single models in our ensemble. In Table~\ref{metrics4} we see that the ensemble clearly has a better sensitivity than each of the single models, meaning it is less prone to making mistakes when predicting the positive class (class of interest). We also see that the F1-score of each individual model hovers around that of the ensemble. Even if we were to compare the metrics of a single model to that of the related study \cite{nagaraj2022developing} we see a large improvement in the F1-score given by our models.

\begin{table}[htbp]
\caption{Comparing median metrics of ensemble of 10 models to single models on evaluating level of knot tying and pushing. We show the seeds for single models in parenthesis.}
\begin{center}
\begin{tabular}{|c|c|c|c|}
\hline
\textbf{Task}&\textbf{Precision}&\textbf{Sensitivity}&\textbf{F1 score}\\
\hline
\textbf{Ensemble}& 0.71  & 0.66 & 0.65 \\ \hline
\textbf{Single (seed=2022)}& 0.73  & 0.64 & 0.68 \\ \hline
\textbf{Single (seed=30548)}& 0.81  & 0.59 & 0.61 \\ \hline
\textbf{Single (seed=85844)}& 0.76  & 0.57 & 0.62 \\ \hline
\textbf{Single (seed=20)}& 0.68  & 0.63 & 0.66 \\ \hline
\textbf{Single (seed=180)}& 0.81  & 0.63 & 0.65 \\ \hline
\textbf{Single (seed=357)}& 0.78  & 0.53 & 0.59 \\ \hline
\textbf{Single (seed=485621)}& 0.81  & 0.63 & 0.64 \\ \hline
\textbf{Single (seed=102314)}& 0.7  & 0.61 & 0.64 \\ \hline
\textbf{Single (seed=305945)}& 0.69  & 0.59 & 0.65 \\ \hline
\textbf{Single (seed=0)}& 0.68  & 0.6 & 0.66 \\ \hline
%\textbf{Average}& 0.68  & 0.6 & 0.66 \\ \hline
%copy& More table copy$^{\mathrm{a}}$& &  \\
%\multicolumn{4}{l}{$^{\mathrm{a}}$Sample of a Table footnote.}
\end{tabular}
\label{metrics4}
\end{center}
\end{table}

\begin{figure}[t!]
\begin{tabular}{c}
\centerline{\includegraphics[scale=0.35]{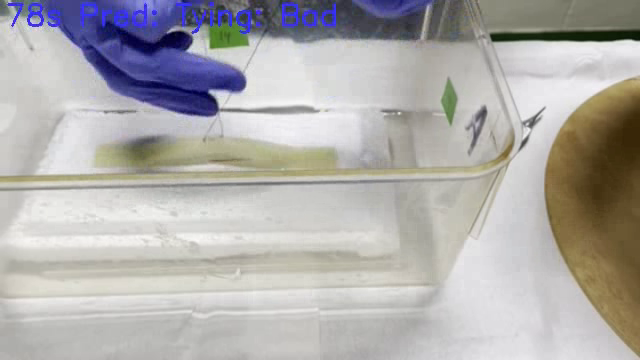}} \\ 
(a) Fumbling and pulling while tying \\
\centerline{\includegraphics[scale=0.35]{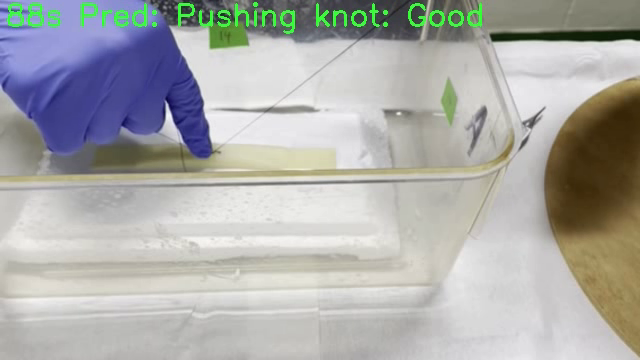}}  \\
(b) Pushing the knot correctly \\
\centerline{\includegraphics[scale=0.35]{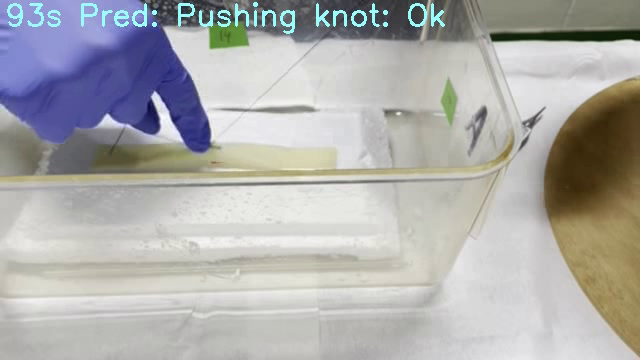}}  \\
(c) Slight pulling while pushing the knot\\
\end{tabular}
\caption{Our model predictions of a video of a fourth year medical student. They have different levels of performance during knot tying which our model captures.}
\label{student2}
\end{figure}

\subsection{Examples of model prediction on a fourth year medical student}
Here we look closely at frameset by frameset prediction of a fourth year medical student video as given by our model. 
The advantage of having a detailed feedback at every frameset is that we can identify specific parts of the 
simulation where the student needs improvement. In Figure~\ref{student2}(a) we see that they fumble and pull while
tying the knot but correctly push the knot later in (b). Both actions and levels are correctly identified by our model and shown in text in the images. Later on as the student is progressing they pull while pushing the knot leading to an okay prediction. In practice such feedback is useful to the student doing the simulation vs a simple yes or no on the entire video (as done elsewhere \cite{nagaraj2022developing}). 

\subsection{Discussion}
While our number of videos is limited in this initial study, our results show that our models can identify the level of knot related tasks better than a random baseline of 50\% and a recent study of the same problem \cite{nagaraj2022developing}. This shows we can potentially create a practical system that would allow medical students and residents to automatically and accurately evaluate and improve their surgical knot tying skills prior to entering the operating room. Such a system could be deployed on a mobile device allowing students to instantaneously get feedback on their performance.

Going forward we plan to add more videos to further boost our model accuracy. Another avenue of future work is to consider the camera view from both horizontal and lateral view. While this may improve the model accuracy, it adds another layer of data that students would need to collect during their exercise. It also requires both horizontal and lateral camera views to be in sync.

\section*{Acknowledgment}
We thank NJIT Academic Research Computing Systems for their computational support.

%\section*{References}

%\bibliographystyle{unsrt}
\bibliographystyle{IEEEtran}
\bibliography{IEEEabrv,my_bib}  %%% Un

\begin{thebibliography}{10}
\providecommand{\url}[1]{#1}
\csname url@samestyle\endcsname
\providecommand{\newblock}{\relax}
\providecommand{\bibinfo}[2]{#2}
\providecommand{\BIBentrySTDinterwordspacing}{\spaceskip=0pt\relax}
\providecommand{\BIBentryALTinterwordstretchfactor}{4}
\providecommand{\BIBentryALTinterwordspacing}{\spaceskip=\fontdimen2\font plus
\BIBentryALTinterwordstretchfactor\fontdimen3\font minus
  \fontdimen4\font\relax}
\providecommand{\BIBforeignlanguage}[2]{{%
\expandafter\ifx\csname l@#1\endcsname\relax
\typeout{** WARNING: IEEEtran.bst: No hyphenation pattern has been}%
\typeout{** loaded for the language `#1'. Using the pattern for}%
\typeout{** the default language instead.}%
\else
\language=\csname l@#1\endcsname
\fi
#2}}
\providecommand{\BIBdecl}{\relax}
\BIBdecl

\bibitem{bridges1999financial}
M.~Bridges and D.~L. Diamond, ``The financial impact of teaching surgical
  residents in the operating room,'' \emph{The American Journal of Surgery},
  vol. 177, no.~1, pp. 28--32, 1999.

\bibitem{allen2016effect}
R.~W. Allen, M.~Pruitt, and K.~M. Taaffe, ``Effect of resident involvement on
  operative time and operating room staffing costs,'' \emph{Journal of Surgical
  Education}, vol.~73, no.~6, pp. 979--985, 2016.

\bibitem{palter2011individualized}
V.~N. Palter and T.~P. Grantcharov, ``Individualized deliberate practice on a
  virtual reality simulator improves technical performance of surgical novices
  in the operating room,'' \emph{Journal of the American College of Surgeons},
  vol. 213, no.~3, p. S126, 2011.

\bibitem{dawe2014systematic}
S.~R. Dawe, G.~Pena, J.~A. Windsor, J.~Broeders, P.~C. Cregan, P.~J. Hewett,
  and G.~J. Maddern, ``Systematic review of skills transfer after surgical
  simulation-based training,'' \emph{Journal of British Surgery}, vol. 101,
  no.~9, pp. 1063--1076, 2014.

\bibitem{zendejas2011simulation}
B.~Zendejas, D.~A. Cook, J.~Bingener, M.~Huebner, W.~F. Dunn, M.~G. Sarr, and
  D.~R. Farley, ``Simulation-based mastery learning improves patient outcomes
  in laparoscopic inguinal hernia repair: a randomized controlled trial,''
  \emph{Annals of surgery}, vol. 254, no.~3, pp. 502--511, 2011.

\bibitem{cox2015moving}
T.~Cox, N.~Seymour, and D.~Stefanidis, ``Moving the needle: simulation’s
  impact on patient outcomes,'' \emph{Surgical Clinics}, vol.~95, no.~4, pp.
  827--838, 2015.

\bibitem{lu2021simulation}
J.~Lu, R.~F. Cuff, and M.~A. Mansour, ``Simulation in surgical education,''
  \emph{The American Journal of Surgery}, vol. 221, no.~3, pp. 509--514, 2021.

\bibitem{nicholas2019simulation}
R.~Nicholas, G.~Humm, K.~MacLeod, S.~Bathla, A.~Horgan, D.~Nally, J.~Glasbey,
  J.~Clements, C.~Fleming, and H.~Mohan, ``Simulation in surgical training:
  Prospective cohort study of access, attitudes and experiences of surgical
  trainees in the uk and ireland,'' \emph{International Journal of Surgery},
  vol.~67, pp. 94--100, 2019.

\bibitem{sharma2020low}
D.~Sharma, V.~Agrawal, J.~Bajajb, and P.~Agarwala, ``Low-cost simulation
  systems for surgical training: a narrative,'' \emph{Journal of Surgical
  Simulation}, vol.~5, pp. 1--20, 2020.

\bibitem{price2011randomized}
J.~Price, V.~Naik, M.~Boodhwani, T.~Brandys, P.~Hendry, and B.-K. Lam, ``A
  randomized evaluation of simulation training on performance of vascular
  anastomosis on a high-fidelity in vivo model: the role of deliberate
  practice,'' \emph{The Journal of thoracic and cardiovascular surgery}, vol.
  142, no.~3, pp. 496--503, 2011.

\bibitem{stefanidis2015simulation}
D.~Stefanidis, N.~Sevdalis, J.~Paige, B.~Zevin, R.~Aggarwal, T.~Grantcharov,
  D.~B. Jones, A.~for Surgical Education Simulation~Committee \emph{et~al.},
  ``Simulation in surgery: what's needed next?'' \emph{Annals of surgery}, vol.
  261, no.~5, pp. 846--853, 2015.

\bibitem{nagaraj2022developing}
M.~B. Nagaraj, B.~Namazi, G.~Sankaranarayanan, and D.~J. Scott, ``Developing
  artificial intelligence models for medical student suturing and knot-tying
  video-based assessment and coaching,'' \emph{Surgical endoscopy}, pp. 1--10,
  2022.

\bibitem{wu2021mvfnet}
W.~Wu, D.~He, T.~Lin, F.~Li, C.~Gan, and E.~Ding, ``Mvfnet: Multi-view fusion
  network for efficient video recognition,'' in \emph{Proceedings of the AAAI
  Conference on Artificial Intelligence}, vol.~35, no.~4, 2021, pp. 2943--2951.

\bibitem{kay2017kinetics}
W.~Kay, J.~Carreira, K.~Simonyan, B.~Zhang, C.~Hillier, S.~Vijayanarasimhan,
  F.~Viola, T.~Green, T.~Back, P.~Natsev \emph{et~al.}, ``The kinetics human
  action video dataset,'' \emph{arXiv preprint arXiv:1705.06950}, 2017.

\bibitem{oord2016wavenet}
A.~v.~d. Oord, S.~Dieleman, H.~Zen, K.~Simonyan, O.~Vinyals, A.~Graves,
  N.~Kalchbrenner, A.~Senior, and K.~Kavukcuoglu, ``Wavenet: A generative model
  for raw audio,'' \emph{arXiv preprint arXiv:1609.03499}, 2016.

\bibitem{bai2018empirical}
S.~Bai, J.~Z. Kolter, and V.~Koltun, ``An empirical evaluation of generic
  convolutional and recurrent networks for sequence modeling,'' \emph{arXiv
  preprint arXiv:1803.01271}, 2018.

\bibitem{picard2021torch}
D.~Picard, ``Torch. manual\_seed (3407) is all you need: On the influence of
  random seeds in deep learning architectures for computer vision,''
  \emph{arXiv preprint arXiv:2109.08203}, 2021.

\bibitem{bethard2022we}
S.~Bethard, ``We need to talk about random seeds,'' \emph{arXiv preprint
  arXiv:2210.13393}, 2022.

\bibitem{godbole2004discriminative}
S.~Godbole and S.~Sarawagi, ``Discriminative methods for multi-labeled
  classification,'' in \emph{Pacific-Asia conference on knowledge discovery and
  data mining}.\hskip 1em plus 0.5em minus 0.4em\relax Springer, 2004, pp.
  22--30.

\bibitem{zhu2004recall}
M.~Zhu, ``Recall, precision and average precision,'' \emph{Department of
  Statistics and Actuarial Science, University of Waterloo, Waterloo}, vol.~2,
  no.~30, p.~6, 2004.

\end{thebibliography}

\end{document}